\documentclass{article}

\usepackage[utf8]{inputenc}
\usepackage{scicite}
\usepackage[margin=1in]{geometry}
\usepackage[T1]{fontenc}
\usepackage{url}
\usepackage{booktabs}
\usepackage{amsfonts}
\usepackage{nicematrix}
\usepackage{graphicx}
\usepackage{pgfplots}
\usepackage{hyperref}
\usepackage{float}
\usepackage{pgfplots}
\pgfplotsset{compat=1.18}
\usepackage{xcolor}

\title{ARC-AGI-3: A New Challenge for Frontier Agentic Intelligence}

\author{ ARC Prize Foundation \thanks{
    \textbf{Development team:}\\
    \textbf{Lead environment designer:} Hunter Henry\\
    \textbf{Engineering:} David Wexler, Derek Smith\\
    \textbf{Environment development:}
    Benjamin Morgan,
    Vadym Andriianov,
    Fraser Scott,
    Pablo Romero Saavedra,
    Jonathan Pappas,
    Flynn Swainston-Calcutt,
    Tom Elliot,
    Kevin Johnson\\
    \textbf{Design and communications:} Bryan Landers\\
    \textbf{President, Board member:} Gregory Kamradt\\
    \textbf{Co-founder, Board member:} Mike Knoop\\
    \textbf{Benchmark designer, Co-founder, Board member:} Fran\c{c}ois Chollet
} }

\begin{document}

\maketitle

\begin{abstract}
We introduce ARC-AGI-3, an interactive benchmark for studying agentic intelligence through novel, abstract, turn-based environments in which agents must explore, infer goals, build internal models of environment dynamics, and plan effective action sequences without explicit instructions. Like its predecessors ARC-AGI-1 and 2, ARC-AGI-3 focuses entirely on evaluating fluid adaptive efficiency on novel tasks, while avoiding language and external knowledge. ARC-AGI-3 environments only leverage Core Knowledge priors and are difficulty-calibrated via extensive testing with human test-takers. Our testing shows humans can solve 100\% of the environments, in contrast to frontier AI systems which, as of March 2026, score below 1\%. In this paper, we present the benchmark design, its efficiency-based scoring framework grounded in human action baselines, and the methodology used to construct, validate, and calibrate the environments.
\end{abstract}

\section{The ARC-AGI benchmark series}

\subsection{ARC-AGI-1 and 2}

In 2019, the Abstraction and Reasoning Corpus (ARC-AGI-1) was introduced alongside the paper ``On the Measure of Intelligence.''~\cite{chollet2019intelligence} The paper proposed a formal framework for evaluating general intelligence as skill-acquisition efficiency rather than task-specific performance. ARC-AGI-1 tests fluid intelligence through grid-based tasks where a pattern must be inferred using limited data. The test taker must discover a novel transformation rule from just a handful of input-output examples. Each task presents pairs of grids of up to 30x30 cells using 10 unique colors, grounded in Core Knowledge priors~\cite{spelke2007} such as objectness and basic geometry. No prior knowledge or previously learned heuristics help in solving ARC-AGI-1.

ARC-AGI-1 was built to resist the memorization-and-retrieval shortcuts that had allowed AI to claim superhuman performance on other tasks like Go~\cite{silver2016alphago}. Each task is unique, which rules out memorization. The small number of examples in each task prevents statistical pattern-matching through massive amounts of training data. This combination made ARC-AGI-1 a durable AI benchmark between 2019 and 2024.

ARC-AGI-2 was introduced in March 2025 to measure complexity scaling of AI reasoning on static tasks. It kept the same grid-based format, while requiring deeper levels of reasoning, featuring multi-step reasoning, sequential rule application, and symbolic interpretation. Every task was human-calibrated with over 400 untrained participants to ensure 100\% solvability. On average, a task from ARC-AGI-1 takes humans about 30 seconds to solve, while a task from ARC-AGI-2 takes humans about 300 seconds to solve.

\subsection{Prior competitions}

The first formal ARC-AGI competition was the 2020 Kaggle Abstraction and Reasoning Challenge~\cite{kaggle2020}, which used ARC-AGI-1. It offered a \$20,000 prize pool and drew 913 teams. The winning solution achieved approximately 20\% accuracy on the private test set using brute-force program search over a library of hand-crafted primitives. This approach would come to define the dominant style of ARC-AGI-1 solvers for the following three years. In 2022 and 2023, Lab42~\cite{arcathon2022} hosted two ``ARCathon'' competitions~\cite{arcathon2023} with \$100,000 in prizes each, expanding international participation.

In 2024, the ARC Prize Foundation~\cite{arcprizefoundation}, co-founded by Mike Knoop and Fran\c{c}ois Chollet, launched the ARC Prize 2024 competition with over \$1 million in prizes. The competition drew 1,430 teams and 47 paper submissions. For the first time, it saw strong performance from deep learning based solutions. Test-time training emerged as a breakthrough technique, reaching a score of 53.5\% on the private ARC-AGI-1 test set.

The ARC Prize 2025 competition ran on the ARC-AGI-2 benchmark, released in March 2025. It drew 1,455 teams and 90 paper submissions. NVIDIA's NVARC team took first place~\cite{nvarc2025} with 24\% accuracy, using synthetic data generation and test-time training on a 4B parameter model. The 85\% grand prize threshold remained unclaimed across both competition years.

\subsection{ARC-AGI-1 and 2 key findings}

\subsubsection{Predictive power}

The Transformer architecture~\cite{vaswani2017attention}, published in 2017, paved the way for the ``scaling era'' of AI, eventually enabling self-supervised LLMs (Large Language Models) for which benchmark performance kept increasing with no further architecture changes, purely by increasing training data and training compute. This is the approach known as \textit{pretraining scaling}, which was the dominant paradigm of AGI research from 2019 to 2024. Chain-of-Thought prompting was discovered in 2022~\cite{wei2022chainofthought} as a way to improve test-time reasoning abilities of LLMs and led to a second scaling paradigm using test-time adaptation (or test-time compute).

ARC-AGI-1 resisted pretraining scaling because base LLMs (without test-time adaptation) are limited to memorization and interpolative retrieval of patterns found in their training data and cannot generalize to never-seen-before tasks, even when these tasks are elementary. The latest generation of base LLMs (as of March 2026) still perform poorly on ARC-AGI-1.

Test-time reasoning was the key innovation that enabled LLM-based systems to begin exhibiting non-zero fluid intelligence, giving rise to the LRM (Large Reasoning Model) paradigm. This was first demonstrated by OpenAI's breakthrough o1 and o3 systems~\cite{chollet2024o3breakthrough} on ARC-AGI-1. It was at the time the only benchmark to precisely identify the advent of frontier AI fluid reasoning.

Modern-day models excel at reasoning, precisely corroborated by ARC-AGI-2 progress. These new capabilities have enabled systems to achieve considerable product-market fit with coding tools (such as Claude Code and Codex). There is now global recognition that AI agents capable of broad task automation will transform software engineering and eventually most of the economy. The capstone achievement of the ARC-AGI-1 and 2 benchmarks was signaling and quantifying the arrival of these transformative advances from late 2024 to late 2025 (see Figure 1).

\begin{figure}[t]
    \centering
    \includegraphics[width=\textwidth]{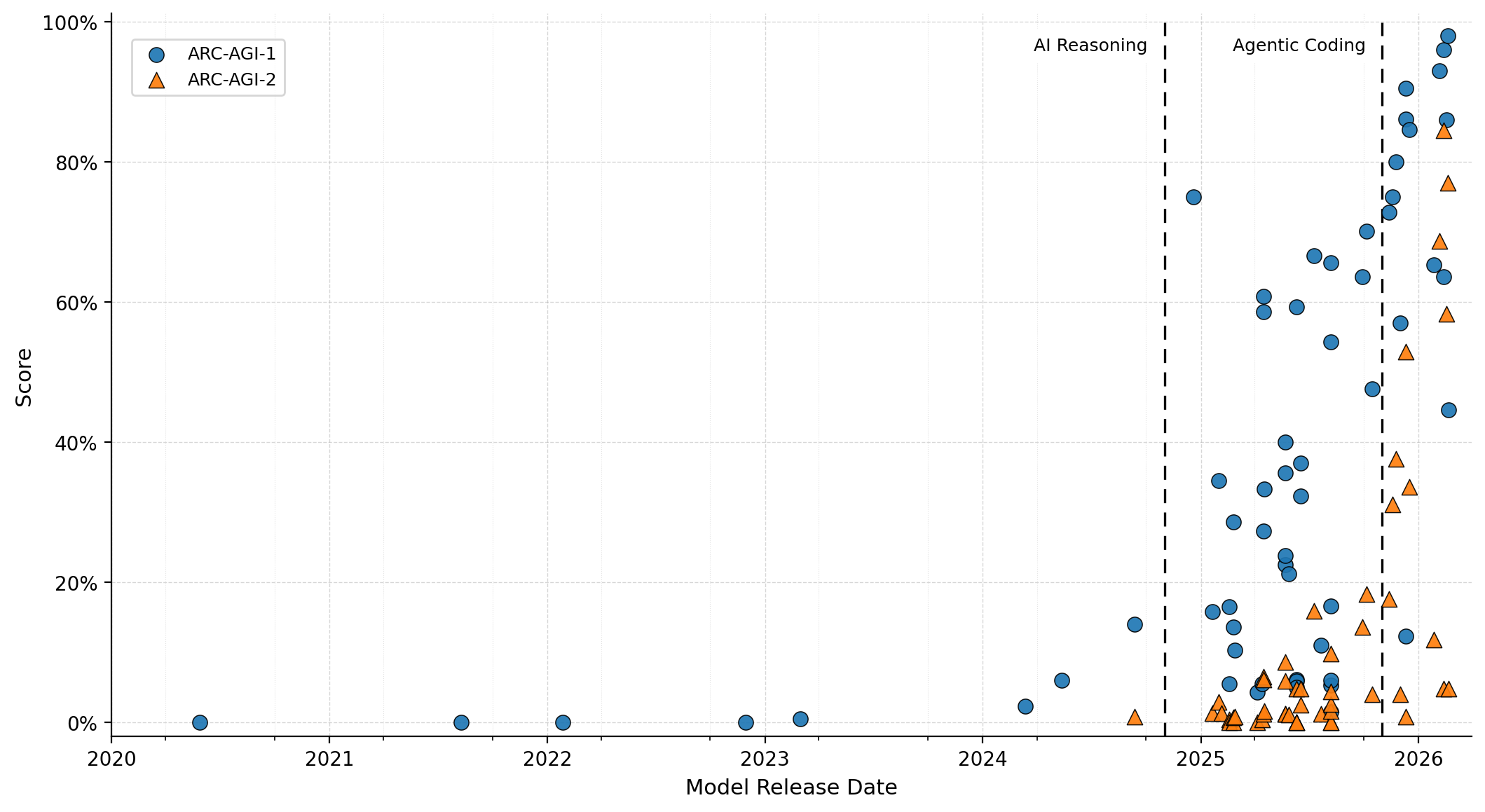}
    \caption{Frontier AI performance on ARC-AGI since introduction in 2019.}
    \label{fig:arc-agi-timeline}
\end{figure}

\subsubsection{Known limits of LRM fluid intelligence}

Modern LRMs enable automation in any domain where the following is true:

\begin{itemize}
    \item The base model contains sufficient knowledge coverage of the domain.
    \item The domain provides an exact correctness feedback measure to the LRM (such domains are known as ``verifiable domains'').
\end{itemize}

This strongly suggests that AI reasoning capability is tied to LRM knowledge. Take a moment to appreciate how strange this is: Human reasoning capability is \textit{not} bound by domain knowledge. This leads to imprecise descriptions of LLMs as ``jagged intelligence'', when in reality LLMs remain bound to task-specific training, albeit now over task-specific \textit{reasoning chains} instead of the literal task data.

Collecting domain knowledge and building verifiers is expensive, as seen with programming environments for AI coding agents. Future LLM automation of new areas will be driven by industry investment, first in domains where it is possible to assemble the necessary data, compute, and labor and where the return on investment is positive. This includes investment to produce new scientific knowledge. In late 2025, Steve Hsu published an example~\cite{hsu2025quantum} of an LRM automation system that discovered a novel result in quantum physics. Many scientific domains, like drug discovery, are highly automatable due to their mechanistic nature.

However, many other potential applications of LRM automation are either too expensive or impractical to deploy today. As model efficiency increases, more applications become possible. But in the bigger picture, machines that can perform highly efficient adaptation to produce paradigm-shifting innovation are still well outside our reach. Even the latest LRMs remain bottlenecked by human intelligence and show limited ability to cover novel domains, which is a key argument why they fall short of AGI.

\subsubsection{Overfitting and memorization shortcuts}

Classically, models are said to be overfit when they learn ``too much'' at training time -- memorizing specific features of training data instances instead of learning general, causal principles that will generalize to new instances. This leads the model to perform well during development (via non-generalizable ``memorization shortcuts''), while performing poorly on production data after deployment, which often comes as a surprise to model developers. With many benchmarks, the test data shares a high degree of similarity with the training data, leading such overfit models to perform well on the test data despite having limited power to generalize to new instances in the wild. This can also happen when the training process leaks information about the test data.

ARC-AGI-1 and ARC-AGI-2 were designed to be resistant to this style of memorization shortcuts, by using a private dataset (inaccessible to model developers) for official scoring and verification and making sure all tasks were reasonably unique -- both distinct from each other and distinct from tasks found on the web.

However, frontier LRMs demonstrate non-zero fluid intelligence and can thus adapt to tasks further away from their training distribution (while still requiring extensive domain knowledge). This means that benchmarks that were designed to resist direct memorization can now be attacked via higher-level shortcuts if the public training set and the private test set are overly similar (e.g., identically distributed) and the model was trained on an enormous amount of tasks representing a dense sampling of the task space -- possibly tasks that were automatically generated for this purpose. A simple strategy to achieve this is to ask a model to generate more tasks from the domain, solve them, verify the solutions via a reliable verifier (to note, ARC-AGI-1 and 2 tasks are verifiable), then train on the produced reasoning traces, in a loop. This can scale to millions of tasks -- enough to provide high coverage density of the target domain at training time, thus considerably reducing or outright removing the need for test-time adaptation.

We believe this has happened to ARC-AGI-1 and ARC-AGI-2 with frontier LRMs -- either incidentally or intentionally. Here is one bit of evidence from our Gemini 3 verification~\cite{gemini3verification}. The model mentions the following in its reasoning chain:

\begin{quote}
\textit{... Target is Green (3). Pattern is Magenta (6) Solid. Result: Magenta Square on Green ...} (Gemini 3 Deep Think)
\end{quote}

Our verification model prompt \textit{does not} mention ``ARC-AGI'' or the integer-to-color mapping used by ARC-AGI tasks, yet the model is using the correct color mapping in its reasoning. This strongly suggests that ARC-AGI data is well represented in the underlying model -- enough to make correct ARC-AGI inferences based on just the structure and format of 2D arrays of integers.

Going forward, benchmark designers will need to steer private datasets to be out-of-distribution (OOD) from any publicly available demonstration data if they want to test true generalization.

\section{ARC-AGI-3}

\subsection{ARC-AGI-3 goals}

The overarching goal of the ARC-AGI series is to measure the ``residual gap'' between current artificial intelligence and human-level AGI. This gap is definitionally a moving target and necessitates new versions as frontier AI capabilities advance. We define AGI not merely as a set of static capabilities, but as a system's ability to acquire any skill a human can, as efficiently as a human can. While ARC-AGI-1 and 2 focused on data-efficient modeling (inferring rules from static input/output pairs), ARC-AGI-3 shifts to targeting \textbf{agentic intelligence}. Specifically, ARC-AGI-3 uses a set of \textbf{interactive turn-based environments} to evaluate a test-taker across four core functional components of agentic intelligence:

\begin{itemize}
    \item \textbf{Exploration:} In real-world environments, information is rarely provided passively. It must be actively obtained by the agent by interacting with its surroundings.
    \item \textbf{Modeling:} Inherited from previous ARC-AGI generations, this is the ability to turn raw observations into a generalizable world model that can predict future states and outcomes.
    \item \textbf{Goal-Setting:} A cornerstone of autonomy, goal-setting is the ability to identify interesting or desirable future states without explicit instructions. The agent must independently determine ``what to target'' based on its own intrinsic drive and environmental cues.
    \item \textbf{Planning and Execution:} This involves the strategic mapping of an action path from the current state to the identified goal. It requires not only initial accuracy but also the agility to course-correct in response to environmental feedback or unexpected results.
\end{itemize}

While causal modeling remains a prerequisite, the benchmark now demands autonomous navigation of ``unknown unknowns''. In particular, one of the most significant hurdles in ARC-AGI-3 is that the agent is \textbf{never told the objective nor provided instructions}. It must autonomously infer the mechanics of each new environment, including the win conditions.

\subsection{Intelligence as efficiency}

In the ARC-AGI-3 framework, intelligence is fundamentally defined as \textbf{efficiency} across the four pillars mentioned above. A high-intelligence system is not simply one that can solve a task, but one that does so while minimizing its resource usage. There are many forms of resources one might consider, such as data (number of states explored), time, compute, and risk (in embodied or game-based scenarios, every action carries a cost: potential ``death,'' loss of progress, or wasted energy). ARC-AGI-3 makes the opinionated decision of subsuming all of these into a single scalar efficiency measure: \textbf{action efficiency}. Action efficiency lets us provide a standardized comparison between biological and artificial agents.

Action efficiency is the number of moves or ``turns'' required to solve a new environment upon first contact with it. It is aggregated on a per-level basis (more in the Scoring Methodology section). This metric has the following useful properties:

\begin{enumerate}
    \item \textbf{It penalizes ``brute-forcing'':} A system that blindly tries many options is viewed as less intelligent than one that quickly forms a model of the environment and uses it for effective planning and execution.
    \item \textbf{It accounts for data efficiency and risk efficiency:} Fewer actions naturally translate to lower exposure to environmental hazards.
    \item \textbf{It enables direct human-AI comparison:} By establishing a baseline of human action efficiency on these same environments, we can quantitatively measure how close an AI is to ``human-level'' skill acquisition.
\end{enumerate}

Beating ARC-AGI-3 is achieved when an AI system matches or exceeds human-level action efficiency on ARC-AGI-3 environments that it sees for the first time, averaged across all private environments.

\subsection{The ARC-AGI-3 environment format}

Each environment in ARC-AGI-3 is structured as a series of levels. A level ends when a win condition (terminal frame) is reached. The benchmark utilizes a turn-based interface designed to prioritize offline reasoning over real-time sensorimotor ``reflexes.'' At each turn, the agent is presented with a frame (or series of frames representing a transition animation), and must take one action to move to the next frame. The environment's state does not change asynchronously from the agent's actions.

\begin{figure}[h]
    \centering
    \includegraphics[width=0.65\textwidth]{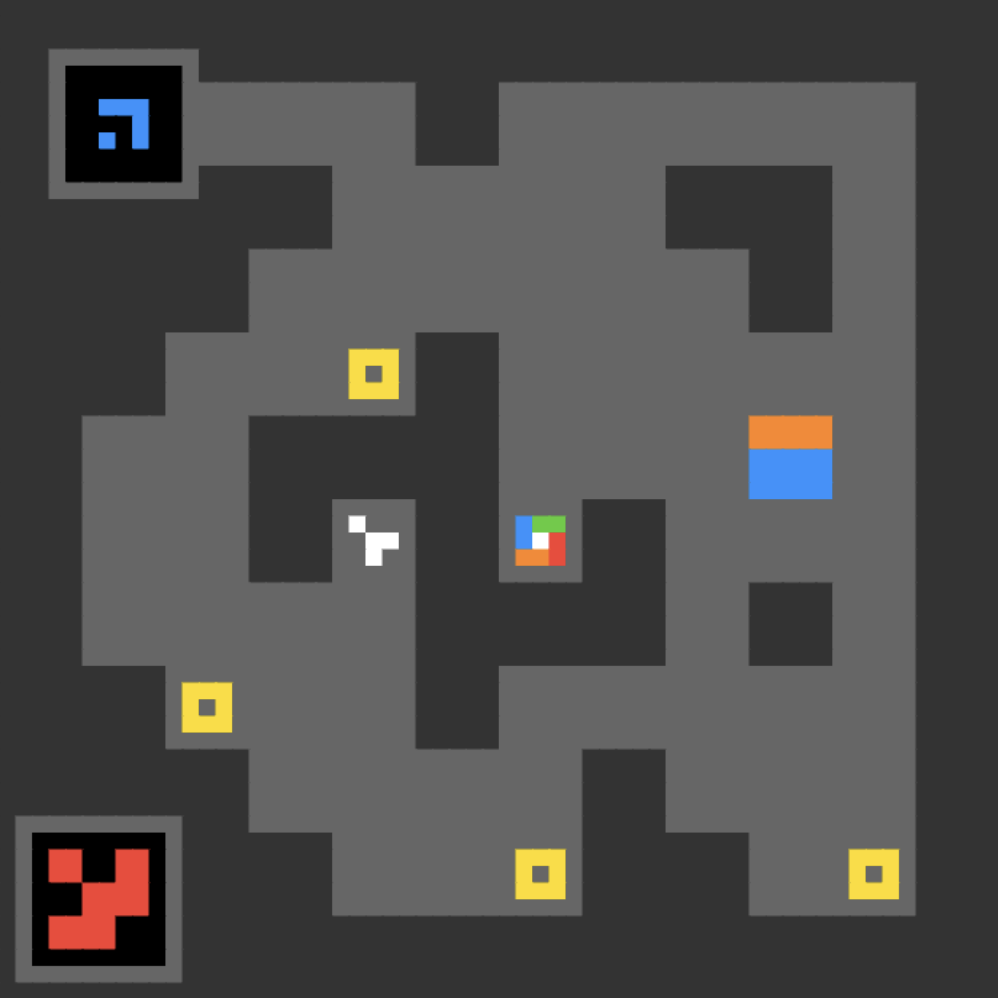}
    \caption{Screenshot of ARC-AGI-3 environment \texttt{ls20}.}
    \label{fig:ls20-screenshot}
\end{figure}

\subsubsection{The Observation Space}

The agent views a \textbf{64x64 grid} where each cell is one of \textbf{16 possible colors}. A given grid state is called a ``frame''. At each turn, the agent receives a frame or frame sequence. Frame sequences allow for non-interactive animations (e.g., an object moving across the screen) between player turns.

\subsubsection{The Action Space}

Each environment offers a different action space, which is a subset of:

\begin{itemize}
    \item Five key actions, plus an Undo action (reverting to the previous state)
    \item One action to select (e.g. click on) a cell from the 64x64 grid by specifying its coordinates
\end{itemize}

This small action space ensures that the complexity of the benchmark lies in the logic of the environment, not the difficulty of the controls.

An action is defined as a discrete interaction with the environment, i.e., a turn where the agent submits a command, move, or input that affects the environment state. Internal operations that do not alter the environment, such as tool calls, reasoning steps, or retries within the model itself, are \textbf{not counted} as actions.

\section{Building ARC-AGI-3}

\subsection{The ARC-AGI-3 game studio}

To create ARC-AGI-3, we established an in-house game studio tasked with producing a collection of novel interactive environments under a shared set of technical and design constraints. This studio model allowed creative environment design to be coupled with standardized interfaces, evaluation procedures, and validation criteria which enabled environment production to scale without sacrificing consistency.

\subsection{Production pipeline}

In order to streamline production, we organized the studio around three core functions: a lead developer who defined and reviewed the production pipeline, individual developers who implemented environments, and an engineer who built high-level automation and internal tools to support creation at scale.

Our initial assumption was that environments could be developed serially by each environment developer, with one environment completed before the next entered production. In practice, this did not reflect the realities of development. Because ideation, implementation, playtesting, and revision progressed at different rates, throughput was highest when three to four environments were in development simultaneously by a single developer, at different stages of the pipeline.

The production pipeline comprised four stages:

\begin{enumerate}
    \item \textbf{Specification:} The developer creates an environment concept description, which is collectively reviewed before implementation, allowing major design issues to be identified early and reducing iteration cost.
    \item \textbf{Internal:} The developer builds a prototype and tests it with members of the team.
    \item \textbf{External:} The environment undergoes external human testing in order to determine whether it satisfied our human-performance criteria. Environments that passed this stage were moved on.
    \item \textbf{Done:} The environment is finished and ready for sorting into one of the ARC-AGI-3 sets.
\end{enumerate}

Throughout the process, automated validation was used to detect development bugs, trivial environments, and regressions before an environment advanced.

\subsection{Technical constraints}

All ARC-AGI-3 environments were implemented in a shared runtime using a custom in-house environment engine. We initially used Unity for this process but found it too heavy and too slow for the rate of iteration required. Building a custom engine provided tighter control over performance, tooling, and evaluation. The final engine is implemented in Python~\cite{arcagi3toolkit} to achieve our minimum performance goal threshold of 1,000 frames per second.

\subsection{Environment design constraints}

The core challenge in ARC-AGI-3 is intended to be reasoning, rather than perception, which is why ARC-AGI-3 is turn-based instead of real-time. Idea generation for new environments, rather than implementation, was often the most challenging part of the development process. Below, we present our core environment design principles.

\textbf{Core knowledge priors only:} To ensure the benchmark remains a test of innate reasoning rather than acquired knowledge, all environments are strictly limited to Core Knowledge priors~\cite{spelke2007}, and seek to avoid similarities with existing games.

\begin{itemize}
    \item \textbf{Objectness:} Elements are perceived as coherent, persistent entities that can move, collide, or be occluded.
    \item \textbf{Basic geometry and topology:} Understanding of symmetries, rotations, and elementary topology (e.g., ``inside'' vs. ``outside'', connectedness, holes).
    \item \textbf{Basic physics:} Intuitive rules like gravity, momentum, and bouncing.
    \item \textbf{Agentness:} Recognizing that certain objects act with intent and pursue goals.
    \item \textbf{No language or cultural symbols:} Environments never use numbers, letters, recognizable real-world clip-art (like flowers or keys), or cultural conventions (like green meaning ``go'').
\end{itemize}

\textbf{Novelty:} Each environment is required to be novel both with respect to preexisting video games, and with respect to the previously created set of environments. As a practical test of novelty, we tested whether a single program could solve two different environments while being at least 50\% shorter than the concatenation of two independent solution programs; when the answer is yes, those environments are likely insufficiently distinct.

\textbf{Human solvable:} Environments are designed to be solvable by humans within a bounded play session of approximately 20 minutes (but most environments can be solved in only a few minutes).

\textbf{Difficulty through composition:} Difficulty is not intended to arise from obscurity or increasing complexity. Rather it is intended to arise from the composition of reasoning demands acquired over the course of play. Later levels are therefore expected to require the accumulation and integration of concepts learned earlier in the environment.

\textbf{Tutorial level:} The first level in an environment functions as a tutorial level and is intentionally easy (to both humans and AI). In some cases, random agents can occasionally stumble into success at this stage, which is acceptable by design. The purpose of the opening level is to communicate the core interaction pattern and orient the player without instructions.

\textbf{Multiple mechanics:} Each environment contains multiple mechanics. Environments centered on a single mechanic that scaled in size or difficulty are treated as an anti-pattern.

\textbf{Levels:} Environments are developed with a level-based structure, with at least six levels per environment.

\textbf{Environment ID:} Each environment has a unique ID that is a series of four characters. Informally, environments have longer names, but these are not shared publicly to avoid sharing semantic information about the environment's goal or mechanics.

\subsection{Automated environment validation}

The validation pipeline is divided into two complementary layers: \textbf{deterministic system qualification} and \textbf{exploratory state-space analysis}. Together, these provided confidence that an environment was both compatible with the platform and behaviorally well-formed under large-scale automated execution.

\subsubsection{Environment qualification}

Qualification focuses on platform integration and reliability. Structural tests verified each environment can be loaded, instantiated, and exercised by the broader runtime environment. Several regimes are deployed.

The first random regime runs for up to 50,000 steps and asserts that no level can be beaten by accident. This is primarily a sanity check against trivial or degenerate reward paths.

The second regime extends to 1,000,000 steps and strengthened the constraint that non-tutorial levels must remain unbeaten under uninformed random play. This helped ensure that genuine progression requires structure rather than luck.

A third 1,000,000-step sweep is run across all levels. This combined performance and fuzzing harness testing. At this scale, random testing becomes useful for surfacing edge-case crashes, malformed transitions, invalid frame outputs, inconsistent hidden-state behavior, and rare action-sequence defects that deterministic tests may miss.

Finally, developer-provided recording-based playback verifies reproducibility. Known-good recordings are replayed under both win and loss conditions, confirming that the engine can serialize and faithfully re-execute action traces. This is important not only for regression testing, but also for debugging, benchmarking, auditability, and future model analysis.

\subsubsection{Exploring graph-based state space construction and win probability estimation}

Exploration models the environment as an explicit directed graph over reachable states (see Figure 3). In this representation, each node corresponds to a unique environment state, while each edge corresponds to a valid player action taken from that state. Node identity is hash-based, allowing the builder to merge distinct trajectories that arrive at the same underlying state. This allows the system to transform repeated simulations into a compact state-space approximation rather than a collection of independent rollouts.

\begin{figure}[h]
    \centering
    \includegraphics[width=0.75\textwidth]{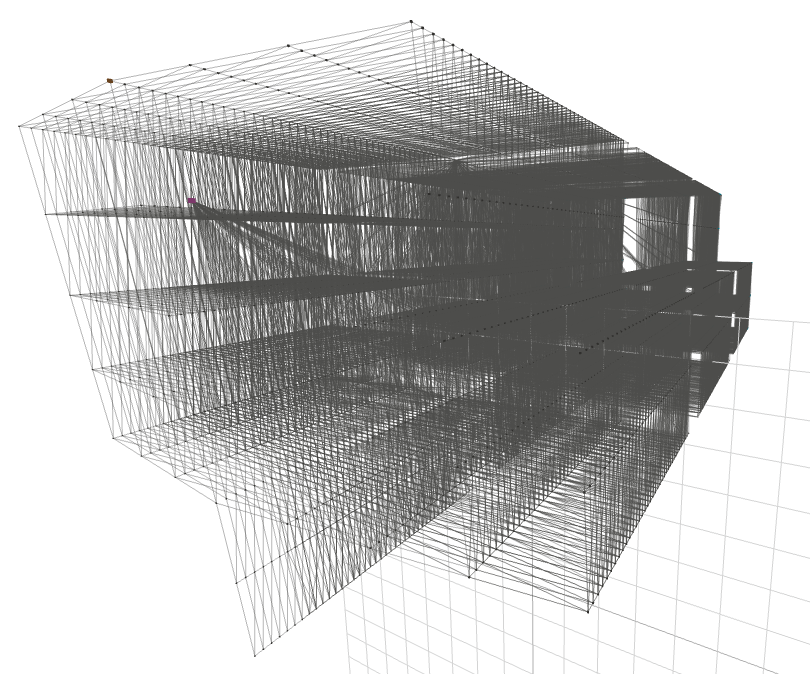}
    \caption{First level of \texttt{ls20} in graph form. Notice the three repeating states -- an artifact of the three-life mechanic of the level. $P_{\text{win}}$ for this level is exactly 1 in 355.}
    \label{fig:graphbuilder}
\end{figure}

Graph construction begins from the reset state of a selected level. For each visited state, the builder enumerates the currently valid actions and records them as outgoing candidate edges. When an edge is followed, the environment advances by one step, a successor node is constructed from the returned frame and hidden state, and that node is either inserted as a new node or merged with an existing equivalent node. Exploration continues until a configured limit is reached, including step budget, time budget, node limit, or edge limit.

Terminal conditions such as level completion and environment completion are explicitly marked. Invalid actions are recorded as transitions that do not change the state. The implementation also tracks merge density, maximum observed depth, cycle detection, and whether the reachable graph has been fully explored.

The graph reveals structural properties of each environment, exposes cycles, estimates reachability, and quantifies stochastic solvability in a reproducible manner. Even when the full graph cannot be exhaustively enumerated, the system can still produce mathematically grounded bounds on the probability that a random policy will solve the environment. Our acceptance threshold was that a random policy should not successfully solve a level more often than 1 in 10,000 times.

\subsection{ARC-AGI-3 environment selection}

The ARC-AGI-3 benchmark consists of the following datasets:

\textbf{Public demonstration set.} The public set is designed to demonstrate the ARC-AGI-3 environment format, while being accessible and engaging for human players. These environments are intentionally easier for both humans and AI, with a stronger emphasis on clarity and fun. As the primary entry point to the benchmark, the public set serves as a community-facing ``front door,'' and is expected to receive the majority of human playthroughs. To preserve evaluation integrity, the public set does not comprehensively represent the mechanics found in the private set, reducing the risk of overfitting or targeted optimization.

\textbf{Private set.} The private set is designed to more rigorously test generalization. These environments are significantly more difficult for both humans and AI, and are intentionally out-of-distribution relative to the public set. They cover a broader and more diverse set of mechanics with limited overlap with the mechanics found in the public environments, and they probe greater adaptation capabilities, involving deeper compositional reasoning (while remaining entirely solvable by humans, which is ensured by our human calibration process).

The private set is further subdivided into two subsets: the \textbf{semi-private set}, to be used to test frontier models behind an external API (therefore subject to a small risk of data leakage), and the \textbf{fully private set}, to be used for the official ARC Prize competition, tightly guarded.

Compared to ARC-AGI-2, which maintains a roughly 10:1 public-to-private ratio, ARC-AGI-3 inverts this balance. The public set shifts from a training resource to a demonstration interface, while the private set becomes the primary basis for evaluation.

\begin{table}[h]
  \centering
  \small
  \begin{tabular}{lp{8cm}r}
    \toprule
    \textbf{Dataset} & \textbf{Purpose} & \textbf{\# of environments} \\
    \midrule
    Public Demo & A public demonstration set that shows the format and basic mechanics of ARC-AGI-3 environments. & 25 \\
    Semi-Private & A private hold out set that is used to test models behind an external API. & 55 \\
    Fully Private & A private hold out that is used for the competition. This set is only given to a very limited number of partners. & 55 \\
    \bottomrule
  \end{tabular}
  \caption{ARC-AGI-3 dataset composition.}
\end{table}

\section{Measuring performance on ARC-AGI-3}

Interactive Reasoning Benchmarks provide a new way to measure the learning efficiency of AI by counting the number of actions (defined below) it takes them to complete a task.

To complete ARC-AGI-3, test takers must use actions in two ways: \textbf{exploration} (learning the environment mechanics and goals) and \textbf{execution} (carry out a strategy to reach a goal) to complete an environment. Counting the total number of actions taken on first exposure to beat an environment accounts for both of these.

\subsection{Scoring methodology}

The ARC-AGI-3 scoring method seeks to score the test taker by its per-level action efficiency (as compared to a human baseline), normalized per environment, across all environments.

This scoring function is called RHAE (Relative Human Action Efficiency), pronounced ``Ray''.

\textbf{The procedure can be summarized as follows:}

\begin{itemize}
    \item \textbf{``Score the AI test taker by its per-level action efficiency''} - For each level that the test taker completes, count the number of actions that it took.
    \item \textbf{``As compared to human baseline''} - For each level that is counted, compare the AI agent's action count to a human baseline, which we define as the upper-median best human action count. For example: If the upper-median best human completed a level in only 10 actions, but the AI agent took 100 to complete it, then the AI agent scores $(10 / 100)^2$ for that level, which gets reported as 1\%. Note that level scoring is calculated using the square of efficiency.
    \item \textbf{``Normalized per environment''} - Each level is scored in isolation. Each individual level will get a score between 0\% (very inefficient) and 115\% (surpasses human level efficiency). The environment score will be a weighted-average of level score across all levels of that environment.
    \item \textbf{``Across all environments''} - The total score will be the sum of individual environment scores divided by the total number of environments. This will be a score between 0\% and 100\%.
\end{itemize}

\textbf{We define the metric as follows:}

For a given level $l$ within an environment $e$, let $a_{l,e}$ be the number of actions taken by the AI agent, and $h_{l,e}$ be the human baseline (defined as the upper-median best human action count).

The \textbf{Level Efficiency Score} $S_{l,e}$ is defined as follows:
\begin{equation}
    S_{l,e} = \min \left( 1.15, \frac{h_{l,e}}{a_{l,e}} \right)^2
\end{equation}

The \textbf{Environment Score} $E_e$ is the linearly weighted average of the level scores, subject to a per-environment cap. For an environment with $n$ levels (where $n=5$), the weight for level $l$ is $w_l = l$. Let $k$ be the number of levels completed by the test taker (where levels are sequential, so completing $k$ levels means completing levels 1 through $k$, and $S_{l,e} = 0$ for uncompleted levels). The score is calculated as:
\begin{equation}
    E_e = \min\left(\frac{\sum_{l=1}^{k} w_l}{\sum_{l=1}^{n} w_l},\; \frac{\sum_{l=1}^{n} w_l \cdot S_{l,e}}{\sum_{l=1}^{n} w_l}\right)
\end{equation}
The first term is the \textbf{environment cap}, which limits the maximum environment score to the weighted fraction of levels completed.

The \textbf{Total Benchmark Score} $T$ is the mean of all environment scores across the dataset $D$:
\begin{equation}
    T = \frac{1}{|D|} \sum_{e \in D} E_e
\end{equation}

This formulation is inspired by the robotics navigation Success weighted by Path Length (SPL) metric~\cite{anderson2018evaluation}, which evaluates not only task completion but also path efficiency.

\begin{figure}[h]
    \centering

\definecolor{pOne}{rgb}   {0.122, 0.467, 0.706}
\definecolor{pTwo}{rgb}   {0.682, 0.780, 0.910}
\definecolor{pThree}{rgb} {1.000, 0.498, 0.055}
\definecolor{pFour}{rgb}  {1.000, 0.733, 0.471}
\definecolor{pFive}{rgb}  {0.173, 0.627, 0.173}
\definecolor{pSix}{rgb}   {0.596, 0.875, 0.541}
\definecolor{pSeven}{rgb} {0.839, 0.153, 0.157}
\definecolor{pEight}{rgb} {1.000, 0.596, 0.588}
\definecolor{pNine}{rgb}  {0.580, 0.404, 0.741}
\definecolor{pTen}{rgb}   {0.773, 0.690, 0.835}
\definecolor{pEleven}{rgb}{0.549, 0.337, 0.294}
\definecolor{medblue}{rgb}{0.290, 0.565, 0.851}  

\pgfplotsset{
  player line/.style={
    line width=0.9pt,
    opacity=0.8,
    no markers,
    sharp plot,          
  },
}

\begin{tikzpicture}

\begin{axis}[
  width=15cm,
  height=8.5cm,
  title={\textbf{ARC-AGI-3 Action Progression (re86)}},
  title style={font=\fontsize{11}{13}\selectfont},
  xlabel={Cumulative Actions},
  ylabel={Score},
  xlabel style={font=\small},
  ylabel style={font=\small},
  xmin=0,  xmax=1700,
  ymin=-0.2, ymax=8.5,
  ytick={0,1,...,8},
  grid=major,
  grid style={gray!20},
  tick label style={font=\footnotesize},
  clip=false,            
]


\addplot[player line, color=pOne] coordinates {
  (0,0)(28,0)(28,1)(66,1)(66,2)(264,2)(264,3)(321,3)
  (321,4)(405,4)(405,5)(522,5)(522,6)(850,6)(850,7)(1071,7)(1071,8)
};
\node[circle, fill=pOne, inner sep=1.8pt] at (axis cs:1071,8) {};
\node[above, font=\fontsize{6}{7}\selectfont\bfseries, text=pOne]
  at (axis cs:1071,8) {100\%};

\addplot[player line, color=pTwo] coordinates {
  (0,0)(24,0)(24,1)(66,1)(66,2)(152,2)(152,3)(212,3)
  (212,4)(401,4)(401,5)(540,5)(540,6)(964,6)(964,7)(1213,7)(1213,8)
};
\node[circle, fill=pTwo, inner sep=1.8pt] at (axis cs:1213,8) {};
\node[above, font=\fontsize{6}{7}\selectfont\bfseries, text=pTwo]
  at (axis cs:1213,8) {100\%};

\addplot[player line, color=pThree] coordinates {
  (0,0)(35,0)(35,1)(76,1)(76,2)(128,2)(128,3)(236,3)
  (236,4)(389,4)(389,5)(470,5)(470,6)(1042,6)(1042,7)(1283,7)(1283,8)
};
\node[circle, fill=pThree, inner sep=1.8pt] at (axis cs:1283,8) {};
\node[above right, font=\fontsize{6}{7}\selectfont\bfseries, text=pThree]
  at (axis cs:1283,8) {100\%};

\addplot[player line, color=pFour] coordinates {
  (0,0)(47,0)(47,1)(93,1)(93,2)(153,2)(153,3)(261,3)
  (261,4)(422,4)(422,5)(589,5)(589,6)(813,6)(813,7)(1033,7)(1033,8)
};
\node[circle, fill=pFour, inner sep=1.8pt] at (axis cs:1033,8) {};
\node[above left, font=\fontsize{6}{7}\selectfont\bfseries, text=pFour]
  at (axis cs:1033,8) {100\%};

\addplot[player line, color=pFive] coordinates {
  (0,0)(26,0)(26,1)(68,1)(68,2)(139,2)(139,3)(391,3)
  (391,4)(558,4)(558,5)(730,5)(730,6)(1202,6)(1202,7)(1470,7)(1470,8)
};
\node[circle, fill=pFive, inner sep=1.8pt] at (axis cs:1470,8) {};
\node[above, font=\fontsize{6}{7}\selectfont\bfseries, text=pFive]
  at (axis cs:1470,8) {89\%};

\addplot[player line, color=pSix] coordinates {
  (0,0)(26,0)(26,1)(66,1)(66,2)(149,2)(149,3)(221,3)
  (221,4)(551,4)(551,5)(679,5)(679,6)
};
\node[circle, fill=pSix, inner sep=1.8pt] at (axis cs:679,6) {};
\node[above, font=\fontsize{6}{7}\selectfont\bfseries, text=pSix]
  at (axis cs:679,6) {56\%};

\addplot[player line, color=pSeven] coordinates {
  (0,0)(20,0)(20,1)(66,1)(66,2)(726,2)(726,3)(939,3)
  (939,4)(1161,4)(1161,5)(1296,5)(1296,6)
};
\node[circle, fill=pSeven, inner sep=1.8pt] at (axis cs:1296,6) {};
\node[above, font=\fontsize{6}{7}\selectfont\bfseries, text=pSeven]
  at (axis cs:1296,6) {44\%};

\addplot[player line, color=pEight] coordinates {
  (0,0)(38,0)(38,1)(99,1)(99,2)(211,2)(211,3)(704,3)
  (704,4)(1047,4)(1047,5)(1598,5)(1598,6)
};
\node[circle, fill=pEight, inner sep=1.8pt] at (axis cs:1598,6) {};
\node[above, font=\fontsize{6}{7}\selectfont\bfseries, text=pEight]
  at (axis cs:1598,6) {26\%};

\addplot[player line, color=pNine] coordinates {
  (0,0)(24,0)(24,1)(62,1)(62,2)
};
\node[circle, fill=pNine, inner sep=1.8pt] at (axis cs:62,2) {};
\node[above, font=\fontsize{6}{7}\selectfont\bfseries, text=pNine]
  at (axis cs:62,2) {8\%};

\addplot[player line, color=pTen] coordinates {
  (0,0)(26,0)(26,1)(66,1)(66,2)
};
\node[circle, fill=pTen, inner sep=1.8pt] at (axis cs:66,2) {};
\node[below left, font=\fontsize{6}{7}\selectfont\bfseries, text=pTen]
  at (axis cs:66,2) {8\%};

\addplot[player line, color=pEleven] coordinates {
  (0,0)(129,0)(129,1)(304,1)(304,2)
};
\node[circle, fill=pEleven, inner sep=1.8pt] at (axis cs:304,2) {};
\node[above, font=\fontsize{6}{7}\selectfont\bfseries, text=pEleven]
  at (axis cs:304,2) {2\%};

\addplot[color=medblue, line width=1.8pt, sharp plot, no markers] coordinates {
  (0,0)(29.9,0)(29.9,1)(78.2,1)(78.2,2)(177.1,2)(177.1,3)(301.3,3)
  (301.3,4)(518.7,4)(518.7,5)(678.5,5)(678.5,6)(1166.1,6)(1166.1,7)(1443.3,7)(1443.3,8)
};
\node[circle, fill=medblue, inner sep=1.8pt] at (axis cs:1443.3,8) {};
\node[right, font=\fontsize{6}{7}\selectfont\bfseries, text=medblue, align=left]
  at (axis cs:1470,7.5) {Median\\actions};

\node[
  anchor=south east,
  font=\fontsize{6.5}{8}\selectfont\ttfamily,
  fill=white,
  fill opacity=0.9,
  text opacity=1,
  draw=gray!50,
  rounded corners=2pt,
  inner sep=4pt,
] at (rel axis cs:0.98,0.02) {%
\begin{tabular}{r@{\hskip 6pt}c@{\hskip 6pt}r}
\textnormal{\textbf{\#}} & \textnormal{\textbf{Lvl/8}} & \textnormal{\textbf{Score}} \\
\hline
\textcolor{pOne}{ 1} & \textcolor{pOne}{8/8} & \textcolor{pOne}{100\%} \\
\textcolor{pTwo}{ 2} & \textcolor{pTwo}{8/8} & \textcolor{pTwo}{100\%} \\
\textcolor{pThree}{ 3} & \textcolor{pThree}{8/8} & \textcolor{pThree}{100\%} \\
\textcolor{pFour}{ 4} & \textcolor{pFour}{8/8} & \textcolor{pFour}{100\%} \\
\textcolor{pFive}{ 5} & \textcolor{pFive}{8/8} & \textcolor{pFive}{89\%} \\
\textcolor{pSix}{ 6} & \textcolor{pSix}{6/8} & \textcolor{pSix}{56\%} \\
\textcolor{pSeven}{ 7} & \textcolor{pSeven}{6/8} & \textcolor{pSeven}{44\%} \\
\textcolor{pEight}{ 8} & \textcolor{pEight}{6/8} & \textcolor{pEight}{26\%} \\
\textcolor{pNine}{ 9} & \textcolor{pNine}{2/8} & \textcolor{pNine}{8\%} \\
\textcolor{pTen}{10} & \textcolor{pTen}{2/8} & \textcolor{pTen}{8\%} \\
\textcolor{pEleven}{11} & \textcolor{pEleven}{2/8} & \textcolor{pEleven}{2\%} \\
\end{tabular}};

\end{axis}
\end{tikzpicture}
    \caption{Action progression and RHAE scoring for environment \texttt{re86}. Each line represents a different human player's playthrough, with the x-axis showing cumulative actions taken and the y-axis showing level reached. Labels indicate the final RHAE score each playthrough achieves.}
    \label{fig:re86-progression}
\end{figure}

\subsection{Key scoring design decisions}

\textbf{Normalize AI scores with upper-median best human score}

A defining characteristic of ARC-AGI-3 scoring is that AI performance gets normalized to human level action efficiency. We conducted in-person human testing in a controlled environment to gather human baseline data. An environment was \textit{only} included in ARC-AGI-3 if it passed an ``easy for humans'' bar. Exactly 10 members of the public are tested on each environment.

We defined the human baseline as the upper-median best human by number of actions used. For a given level, we rank all participants who completed that level by action count and select the upper-median performer (e.g., the 3rd-place finisher among 4 or 5 completions). This provides a robust human capability baseline that is resistant to outlier performance while reflecting strong but representative human efficiency.

\textbf{Per-Level vs Per-Environment aggregation}

AI performance is compared to human performance \textit{per-level}, then aggregated per environment.

Per-level aggregation strips away noise from uneven level lengths. For instance, we've observed that \texttt{vc33}'s level 6 requires 10x actions of its level 1 (50 vs $<$5). A 10\% decrease in efficiency on \texttt{vc33} level 6 (+5 actions) would drown out the efficiency observed on level 1.

If we disregard level efficiency and simply look at actions across an entire environment, the longest levels dominate the score and reduce signal from short levels.

Additionally, early levels are meant to be trivial, late levels are more difficult. If we collapse scoring into one environment-denominator you would not be able to tell \textit{where} the agent is weak. With per-level scores you immediately see ``agent matches human efficiency on levels 1 through 3, but does not perform on levels 4 and 5''.

Lastly, by reinforcing efficiency per level, we won't design environments (or encourage AI) to waste actions on levels because they're still ``under budget'' for a given environment.

\textbf{Cap the maximum per-level score}

To stop a single technical anomaly or outlier level from distorting an entire environment score, we cap the per-level efficiency an AI can receive at 1.15x human baseline.

For example, suppose the human baseline shows 20 actions needed to complete a level, but an AI discovers a 2-action exploit, the ratio (20/2 = 10x) would overwhelm the environment average. To counter this, we cap the maximum score for a level at 1.15x the human baseline. The 1.15 cap allows AI systems that are moderately more efficient than the human baseline to receive credit for that efficiency, while still preventing extreme outliers from distorting environment scores.

\textbf{Power law scoring}

ARC-AGI-3 uses a power-law efficiency term rather than a linear one. For each level, a test-taker's efficiency is defined relative to the human baseline, and that value is then squared before contributing to the final score. This adjustment increases the penalty for highly inefficient solutions while preserving partial credit. Under a linear formulation, substantial inefficiencies can still yield disproportionately high scores (e.g., 2× the human action count yields 50\% credit), reducing the metric’s ability to distinguish between near-human and materially suboptimal performance. The power-law transformation improves this discrimination by more heavily penalizing deviations from the human baseline.

For example, if a human completes a level in 10 actions and an AI system requires 100 actions, the raw efficiency is $10/100 = 0.1$. Under the power-law formulation, this becomes $0.1^2 = 0.01$, or 1\% credit for that level.

\textbf{Weighted levels}

ARC-AGI-3 also applies level weighting within each environment. Because the earliest levels are intentionally easier and in some cases may be solvable through limited exploration or even chance, they should contribute less to the overall environment score than later levels. We therefore use a simple linearly weighted average across the five levels of an environment.

For example, in a 5 level environment, the per-level score contribution is as follows:

\begin{itemize}
    \item Level 1 contributes 1/15th of the environment score
    \item Level 2 contributes 2/15th of the environment score
    \item Level 3 contributes 3/15th of the environment score
    \item Level 4 contributes 4/15th of the environment score
    \item Level 5 contributes 5/15th of the environment score
\end{itemize}

This ensures that introductory or tutorial-like levels have the smallest influence on the final score, while later levels, which typically require a more complete understanding of the environment's mechanics, contribute the most.

\textbf{Per-environment cap}

Because the per-level score can exceed 1.0 (up to 1.15x), we introduce a per-environment score cap to prevent the score from being inflated by high efficiency on a subset of levels. The per-environment cap limits the maximum environment score to the weighted fraction of levels completed. For example, in a 5-level environment:

\begin{itemize}
    \item Completing all 5 levels: environment cap is $15/15 = 100\%$
    \item Completing 4 out of 5 levels: environment cap is $10/15 \approx 66.7\%$
    \item Completing 3 out of 5 levels: environment cap is $6/15 = 40\%$
\end{itemize}

This ensures that completing more levels is always rewarded and that an agent cannot achieve a disproportionately high environment score by being extremely efficient on only a few levels while failing to complete later, more difficult levels.

\subsection{Leaderboards}

ARC-AGI-3 scores are reported on a similar 2D plot as the prior ARC Prize leaderboard used for ARC-AGI-1 and 2. The Y-Axis represents performance (action efficiency, as defined above). The X-Axis continues to represent cost for a given run.

Given the computational cost of evaluating high-reasoning frontier models across the full ARC-AGI-3 evaluation set (tens of thousands of dollars in API costs) we impose an action budget of five times the human-baseline median action count per level. That is, for a level with a human median of n actions to completion, the agent is terminated after 5n actions. This constraint may yield marginally lower reported scores than would be obtained under the environment's intrinsic action limit. However, given the power-law decay of the scoring function, the resulting differential is negligible.

Score reporting from the ARC Prize Foundation will be split into two different leaderboards: an official leaderboard, and a community leaderboard.

\subsubsection{Official leaderboard}

Our intent with the official leaderboard is to accurately help the public sense how close frontier models are to human-level general intelligence. We see general intelligence as the ability to deal with problems that the system was not specifically designed or trained for. This means that the official leaderboard will seek to discount score increases that come from direct targeting of ARC-AGI-3, to the extent possible.

We seek to fight two forms of overfitting that would muddy public sensefinding:

\textbf{Task-specific overfitting.} This includes any agent that is created with knowledge of public ARC-AGI-3 environments, subsequently being evaluated on the same environments. It could be either directly trained on these environments, or using a harness that is handcrafted or specifically configured by someone with knowledge of the public environments. We know such agents can in principle achieve a 100\% score on the public set. To demonstrate it, we are releasing an open-source ``harness'' which scores 100\% on all public environments, using human replay. Because it is impossible to ensure that system designers don't use the public environments as part of their work, and because the public set is materially easier than the private set, we will never report public set scores of any system on the official leaderboard. The public set is to be used strictly as a demonstration of what ARC-AGI-3 is -- evaluating on it is emphatically \textit{not} a valid measure of progress towards AGI.

\textbf{Domain-specific overfitting.} This includes any agent that is created specifically to play ARC-AGI-3 environments in general, either by being trained on many synthetically generated ARC-AGI-3 lookalike environments, or using a harness that contains ARC-AGI-3 specific strategies. We know that by injecting a high amount of human instructions into a harness, or even hand-crafting harness configuration choices such as which tools to use, it is possible to artificially increase performance on ARC-AGI-3 (without improving performance on any other domain). The purpose of ARC-AGI-3 is not to measure the amount of human intelligence that went into designing an ARC-AGI-3 specific system, but rather to measure the general intelligence of frontier AI systems.

Therefore, \textbf{we will focus on reporting the performance of systems that have not been specially prepared for ARC-AGI-3, served behind a general-purpose API} (representing \textit{developer-aware generalization} on a new domain as per~\cite{chollet2019intelligence}). This is similar to looking at the performance of a human test-taker walking into our testing center for the first time, with no prior knowledge of ARC-AGI-3. We know such test takers can indeed solve ARC-AGI-3 environments upon first contact, without prior training, without being briefed on solving strategies, and without using external tools.

To measure this, \textbf{the official leaderboard will not use a harness to report official scores}. Our position is that future AGI systems will not need task-specific external handholding to approach new tasks.

This approach is further justified by our early testing. We hired researchers to build general harnesses targeting a small set of public environments: \texttt{ls20}, \texttt{ft09}, and \texttt{vc33}. We then tested the harnesses on the full public set (which researchers did not have access to at the time). We found extreme bimodal performance across the two sets, controlling for the same frontier model. For example, in a variant of environment TR87, Opus 4.6 scores 0.0\% with no harness and 97.1\% with the Duke harness~\cite{fox2026hillclimbing_arcagi3}, yet in environment BP35, Opus 4.6 scores 0.0\% under both configurations. This is clear evidence that:

\begin{itemize}
    \item Frame content perception and API format are not limiting factors for frontier model performance on ARC-AGI-3: with the right handcrafted strategy, frontier models can in fact solve such environments via the current API format.
    \item Specifically engineered harnesses are not a useful way to measure AGI progress, as their performance on seen environments does not translate to unseen environments, much less to novel domains.
\end{itemize}

In order to make apples-to-apples comparisons across all different models, we will be using the same system prompt for all evaluation runs. Models will not be given tools (although they could be using their own tools behind the model's API, which is a blackbox). The code used in the community leaderboard can be found on Github~\cite{arcagi3communityleaderboard}.

ARC-AGI-3 system prompt:

\begin{quote}
\textit{``You are playing a game. Your goal is to win. Reply with the exact action you want to take. The final action in your reply will be executed next turn. Your entire reply will be carried to the next turn.''}
\end{quote}

At release, frontier models score on the official ARC-AGI-3 leaderboard as follows:

\begin{table}[h]
  \centering
  \small
  \begin{tabular}{llr}
    \toprule
    \textbf{Provider} & \textbf{Model} & \textbf{Score} \\
    \midrule
    Anthropic & Opus 4.6 (Max) & 0.50\% \\
    Google & Gemini 3.1 Pro Preview & 0.40\% \\
    OpenAI & GPT 5.4 (High) & 0.20\% \\
    xAI & Grok-4.20 (Beta 0309 Reasoning) & 0.10\% \\
    \bottomrule
  \end{tabular}
  \caption{Semi-private leaderboard scores for frontier models at release.}
\end{table}

\subsubsection{Community leaderboard}

While the official leaderboard does not include scores achieved using domain-specific harnesses, we recognize the importance of harness research. Better harnesses lead to further task automation abilities, which, while not necessarily representing progress towards AGI, remains economically valuable. We expect that 2026 will see significant progress on harness innovation due to ARC-AGI-3.

To provide a dedicated venue for harness-driven results and to highlight this class of innovation, we introduce a secondary leaderboard, the \textit{community leaderboard}. This leaderboard will be public. Anybody can submit to it, and scores will be self-reported. By default, the ARC Prize foundation will not verify anything on the community leaderboard, and we specifically caution against interpreting any scores on the community leaderboard as evidence of AGI progress.

To note, we expect that the best ideas originating from harness research, if they are sufficiently general, will end up flowing behind the model API layer. For example, the original chain-of-thought research started out as a third-party harness from DeepMind wrapping GPT-3. This evolved into a first-party harness internally at OpenAI (Q* and then Strawberry) and OpenAI later brought this innovation to market as a first-party model called o1. We expect this trend to continue. The best and most general ideas will flow from independent third-party research into first-party harnesses and ultimately into first-party models.

\section{Human calibration and solvability}

In order for an environment to be included in ARC-AGI-3, it needs to pass the minimum ``easy for humans'' threshold. Each environment was attempted by 10 people. Only environments that could be \textbf{fully solved by at least two human participants} (independently) were considered for inclusion in the public, semi-private and fully-private sets. Many environments were solved by six or more people. As a reminder, an environment is considered solved only if the test taker was able to complete all levels, upon seeing the environment for the very first time.

As such, \textbf{all ARC-AGI-3 environments are verified to be 100\% solvable by humans with no prior task-specific training}.

If an environment did not meet the minimum solvability threshold, it is returned to the environment developer for iteration. To diagnose failure modes, we analyze participant performance at the level of individual environments and levels. In particular, we examine per-level completion rates across participants to identify consistent drop-off points, which often indicate unclear mechanics or unintended difficulty spikes.

To further understand the human performance issues, we review full video replays for each level. These replays provide step-by-step visibility into participant behavior, making it possible to observe precisely where and how test takers become stuck. In practice, this combination of aggregate level statistics and detailed replay analysis enables easy identification and correction of problematic mechanics, ensuring that environments meet the intended ``easy for humans'' standard.

\subsection{Testing protocol}

In ARC-AGI-2, human evaluation was conducted in a batch setting, with large-scale testing sessions involving hundreds of participants two to three months apart. In contrast, ARC-AGI-3 adopts a continuous evaluation model to support faster development. Rather than infrequent large cohorts, we conducted smaller-scale testing sessions multiple times per week (Monday, Wednesday, and Friday) at a dedicated testing center in San Francisco. This shift enabled faster feedback cycles.

Participants were presented with a sequence of candidate environments and asked to solve them to the best of their ability within a 90-minute session. No task-specific instructions were provided. Each environment was subject to a soft time limit of 20 minutes, after which participants were prompted to conclude their attempt, and a hard cutoff of 30 minutes was enforced.

Participants received a fixed participation fee of \$115–\$140 for completing the session, along with a \$5 performance-based incentive for each environment successfully solved. This incentive structure was designed to encourage completion while maintaining consistent engagement across environments.

On average, participants completed approximately nine environments per session. A small subset of participants exhibited low-effort behavior, rapidly cycling through environments with minimal engagement. They often abandoned more difficult tasks in favor of attempting a larger number of easier ones. These sessions were excluded from analysis, as this behavior appeared to stem from a misinterpretation of the incentive structure.

An ``attempt'' was defined as a play session consisting of more than 30 actions and less than 30 minutes of interaction. Participants were limited to a single attempt per environment and could not revisit previously completed levels. However, they were allowed to reset the current level at any time. In some cases, participants reset levels after reaching a solution in order to improve efficiency, though this typically increased total interaction time.

\subsection{Participant demographics}

Study participants came from diverse professional backgrounds. Participants were members of the general public and not selected for any special training, abilities, or skill sets (see Figure 4).

\begin{figure}[h]
    \centering
    \includegraphics[width=1\textwidth]{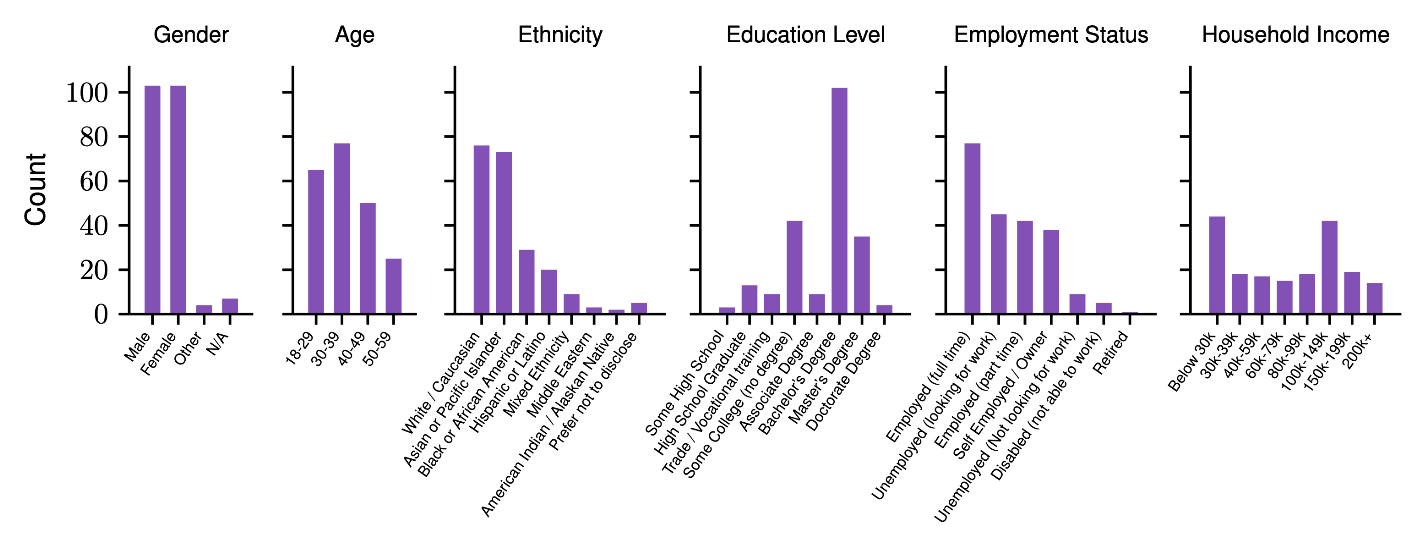}
    \caption{Participant demographics.}
    \label{fig:participant-demographics}
\end{figure}

\subsection{Human performance on ARC-AGI-3}

In total, we recorded 486 unique participants across 414 candidate environments. This resulted in 2,893 total environment attempts.

\begin{figure}[H]
    \centering

\definecolor{histblue}{rgb}{0.290, 0.565, 0.851}  

\begin{tikzpicture}

\begin{axis}[
  width=15cm,
  height=8cm,
  title={\textbf{Per-Level Efficiency Distribution Relative to Human Baseline}},
  title style={font=\fontsize{11}{14}\selectfont, align=center},
  xlabel={Efficiency vs Median},
  ylabel={Count (level completions)},
  xlabel style={font=\small},
  ylabel style={font=\small},
  xmin=0, xmax=310,
  ymin=0, ymax=270,
  xtick={0,50,100,150,200,250,300},
  xticklabels={0\%,50\%,100\%,150\%,200\%,250\%,300\%},
  tick label style={font=\footnotesize},
  clip=false,
  ybar,
  bar width=5.84,
  bar shift=2.92,
]

\addplot[
  fill=histblue,
  fill opacity=0.8,
  draw=white,
  line width=0.4pt,
] coordinates {
  (8.18, 9)
  (14.01, 16)
  (19.85, 22)
  (25.69, 26)
  (31.52, 37)
  (37.36, 21)
  (43.19, 38)
  (49.03, 48)
  (54.87, 43)
  (60.70, 46)
  (66.54, 53)
  (72.38, 68)
  (78.21, 55)
  (84.05, 71)
  (89.89, 76)
  (95.72, 252)
  (101.56, 79)
  (107.40, 66)
  (113.23, 43)
  (119.07, 52)
  (124.91, 45)
  (130.74, 43)
  (136.58, 39)
  (142.42, 39)
  (148.25, 33)
  (154.09, 25)
  (159.92, 26)
  (165.76, 19)
  (171.60, 18)
  (177.43, 24)
  (183.27, 18)
  (189.11, 14)
  (194.94, 19)
  (200.78, 9)
  (206.62, 5)
  (212.45, 9)
  (218.29, 7)
  (224.13, 9)
  (229.96, 4)
  (235.80, 2)
  (241.64, 10)
  (247.47, 4)
  (253.31, 3)
  (259.14, 6)
  (264.98, 5)
  (270.82, 2)
  (276.65, 2)
  (282.49, 4)
  (288.33, 1)
  (294.16, 49)
};

\draw[red, dashed, line width=1.5pt]
  (axis cs:100,0) -- (axis cs:100,265);
\node[
  anchor=north east,
  font=\footnotesize,
  fill=white,
  fill opacity=0.9,
  text opacity=1,
  draw=gray!50,
  rounded corners=2pt,
  inner sep=4pt,
] at (rel axis cs:0.98,0.97) {%
  \tikz\draw[red, dash pattern=on 3pt off 2pt, line width=1.5pt] (0,0) -- (1cm,0);
  \hspace{3pt}Matches median (100\%)%
};

\end{axis}
\end{tikzpicture}
    \caption{Per-level efficiency distribution relative to the median human baseline across all public environments with human data (n=1614 level completions across 340 sessions). The dashed red line marks 100\% (median efficiency).}
    \label{fig:efficiency-histogram}
\end{figure}

\subsubsection{Duration}

The total recorded play time for all attempts was 427.9 hours. The median duration of an attempt was 7.4 minutes (see Figure 5). Successful attempts had a median duration of 8.1 minutes, whereas unsuccessful attempts had a median of 5.9 minutes. Unsuccessful attempts under 20 minutes had a median of 4.7 minutes.

\begin{figure}[h]
    \centering
    \includegraphics[width=0.5\textwidth]{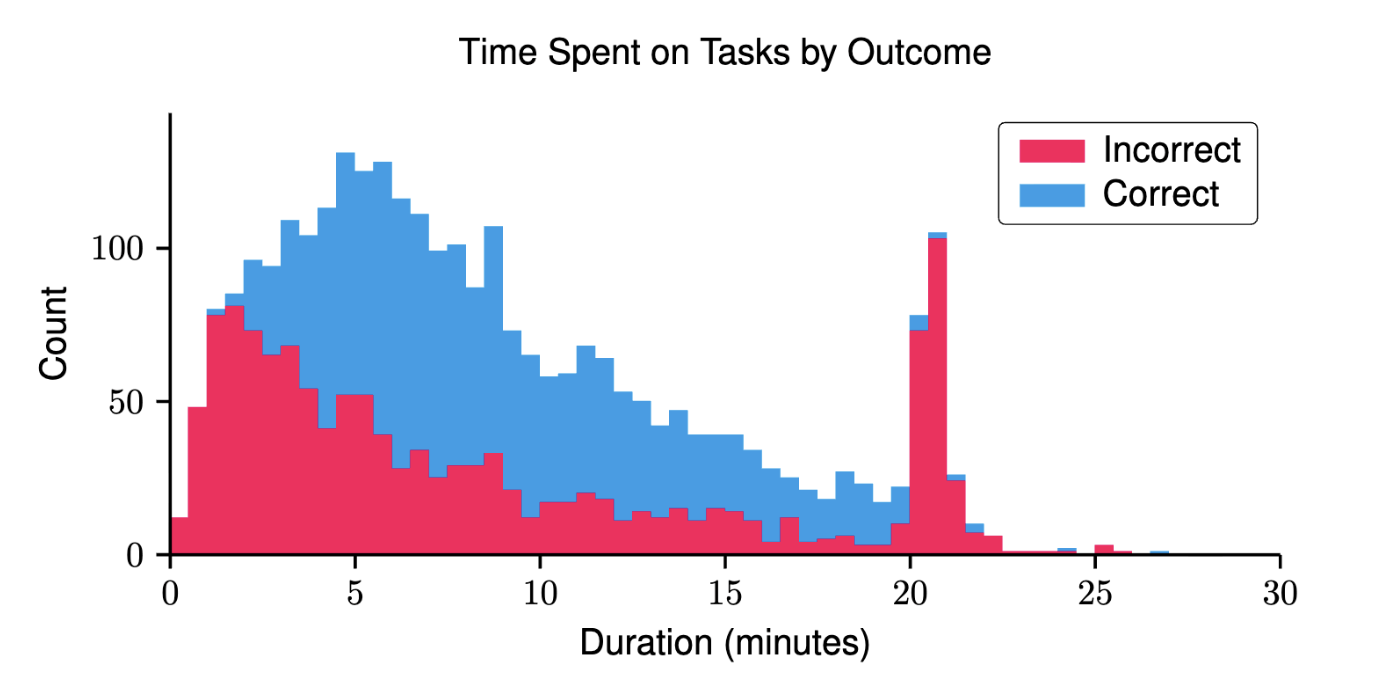}
    \caption{Time spent on environments by outcome, split between successful runs (``correct'') and unsuccessful runs (``correct'').}
    \label{fig:time-by-outcome}
\end{figure}

\subsubsection{Human efficiency}

The human efficiency of beating ARC-AGI-3 is measured by the number of actions it took to complete the environment. Because all human evaluations were conducted as first-run attempts, this data allows us to measure how efficiently humans solve each environment when encountering it for the first time. We track three reference points (see Figure 6):

\begin{itemize}
    \item Optimal playthrough: Empirical estimate of the lower bound on the number of actions needed to solve the environment (once the environment's mechanics and goals are already fully understood.)
    \item Best first-run playthrough: Best first-run human playthrough aggregated per level. It combines the fewest actions achieved by any test participant on each individual level on a first run, regardless of whether they came from the same person.
    \item Human baseline: Upper-median best first-run human playthrough. This is what we use as the human baseline in the official score computation.
\end{itemize}

The difference between the ``optimal playthrough'' and the ``best first-run playthrough'' captures the amount of actions that need to be expended for initial exploration and mechanics learning.

\begin{figure}[H]
    \centering
    \begin{tikzpicture}
    \begin{axis}[
        width=13cm,
        height=6cm,
        title={Total Actions by Level: Environment \texttt{ls20}},
        xlabel={Total \# Actions},
        ylabel={Level},
        xmin=0, xmax=800,
        ymin=0, ymax=7,
        xtick={0,100,200,300,400,500,600,700,800},
        ytick={0,1,2,3,4,5,6,7},
        grid=both,
        major grid style={gray!60},
        minor grid style={gray!25},
        legend style={
            at={(0.98,0.02)},
            anchor=south east,
            fill=white,
            draw=gray!50
        },
        legend cell align={left},
        tick align=outside,
    ]

    \addplot[thick, blue] coordinates {
        (0,0) (13,1) (58,2) (97,3) (140,4) (184,5) (256,6) (309,7)
    };
    \addlegendentry{Optimal Playthrough}

    \addplot[thick, orange] coordinates {
        (0,0) (14,1) (59,2) (98,3) (149,4) (203,5) (289,6) (395,7)
    };
    \addlegendentry{Best First Run Playthrough}

    \addplot[thick, green!60!black] coordinates {
        (0,0) (21,1) (144,2) (183,3) (275,4) (329,5) (437,6) (546,7)
    };
    \addlegendentry{Human Baseline}

    \end{axis}
    \end{tikzpicture}
    \caption{Total actions by level for environment \texttt{ls20}.}
    \label{fig:ls20-actions}
\end{figure}

\section{ARC-AGI-3 pre-launch testing}

Unlike ARC-AGI-1 and 2, we decided to release previews of ARC-AGI-3 prior to the full launch in order to guide our final benchmark design. This gave us critical feedback on what environments were easier and more engaging, and enabled early AI tests to vet our design choices.

To incentivize this, we both hosted an agent preview competition~\cite{kamradt2025arcagi3preview} and worked with external teams to red-team ARC-AGI-3.

\subsection{Agent Preview Competition}

The ARC-AGI-3 Preview Agent Competition~\cite{kamradt2025arcagi3preview} ran for 30 days from July 18 to August 19, 2025. Three public environments were released and three private environments were held back as a hidden evaluation set. Final scoring only measured the ability for AI systems to generalize to this hidden evaluation set.

Top entries included:

\begin{itemize}
    \item StochasticGoose, Tufa Labs (12.58\%, first place)~\cite{smit2025stochasticgoose}: A CNN with reinforcement learning to predict which actions would cause frame changes, encoding 64x64 frames through a four-layer convolutional network. It achieved 12.58\% and completed 18 levels.
    \item Blind Squirrel (6.71\%, second place)~\cite{blindsquirrel2025}: A directed state graph from observed frames.
\end{itemize}

Both winning approaches used an informed search approach, exploring as much of the action space of the environment as possible in the hope of encountering a winning combination by chance.

\subsection{ARC-AGI-3 academic partners}

To complement internal development, ARC-AGI-3 included a small number of academic partnerships aimed at exploring agentic approaches to ARC-AGI-3. These collaborations provided early signal on how frontier models behave on the benchmark while helping surface key challenges in harness design, particularly around context management and long-horizon reasoning.

One of the collaborations was with Duke University, which participated through a sponsored effort led by a small research team. Their work focused on building an agentic harness around a large reasoning model (LRM), with particular emphasis on managing interaction history and extracting relevant state from prior actions.

Context management is a central challenge in ARC-AGI-3. Environment frames are 64x64 grids, and maintaining a naive rolling window of observations quickly exhausts a model's context budget. Their harness~\cite{fox2026hillclimbing_arcagi3} addresses this by allowing the model to execute arbitrary Python code to selectively retrieve and transform information from its action history. This enables more targeted reasoning over past states and improves decision-making efficiency. In evaluation, this approach was able to solve all three public environments with action counts comparable to human performance.

\subsection{Community approaches and early experimentation}

In parallel with academic partnerships, ARC-AGI-3 was released early to the broader research community to encourage independent experimentation. This early access surfaced a range of novel harness designs and provided additional insight into the emerging design space of agentic systems.

Symbolica AI introduced a harness called \textit{Arcgentica}~\cite{symbolica2026arcgentica}, which employs an orchestrator–subagent architecture. A top-level orchestrator does not interact with the environment directly. Instead, it delegates tasks to specialized subagents that return compressed textual summaries. This design constrains context growth and allows the system to maintain a higher-level plan without exceeding context limits. This approach was also able to solve all three public environments.

\section{ARC Prize 2026}

A key part of the ARC Prize Foundation is hosting the annual ARC Prize competition. This will continue in 2026 across two tracks: The ARC-AGI-3 track and the ARC-AGI-2 track.

The total prize pool is now \$2M to encourage open research. Both competitions are held on Kaggle. As always, participants must open source their solutions in order to receive prize money. This will be the final year of the ARC-AGI-2 track, and as such the grand prize is guaranteed to be paid out to the best team this year. The primary focus going forward will be on ARC-AGI-3.

Previous year competition recaps for 2024~\cite{arcprize2024recap} and 2025~\cite{arcprize2025recap} can be found at \href{https://arcprize.org}{arcprize.org}.

\section{Conclusions}

ARC-AGI-3 introduces an interactive reasoning benchmark for evaluating agentic intelligence, focusing on a system's efficiency at acquiring new skills through exploration, model formation, goal inference, and planning in unfamiliar environments. By structuring environments around core knowledge priors and measuring performance through action efficiency relative to human baselines, the benchmark aims to capture aspects of general intelligence that are not reflected in static evaluation settings. The transition to interactive environments provides a more direct test of whether systems can adapt to ``unknown unknowns'' without reliance on prior exposure or task-specific optimization.

The development of ARC-AGI-3 also highlights the importance of benchmark design in an era of increasingly capable models. Lessons from earlier ARC-AGI benchmarks suggest that even carefully constructed static datasets can become susceptible to overfitting as training data expands to directly target the benchmark. In response, ARC-AGI-3 emphasizes novelty, compositional generalization, and out-of-distribution design, along with human calibration, to maintain a meaningful evaluation signal.

Initial results from human studies and early AI systems suggest that ARC-AGI-3 presents a qualitatively different challenge from prior benchmarks. Humans are able to reliably solve the environments within bounded time and action budgets, while current AI systems struggle to achieve consistent performance without significant external scaffolding. This gap reflects not only differences in reasoning capability, but also limitations in exploration strategies, hypothesis revision, and efficient planning under uncertainty.

To our knowledge, ARC-AGI-3 is the only unsaturated general agentic intelligence benchmark as of March 2026.

As models continue to improve, evaluation frameworks must evolve to remain informative and resistant to shortcut solutions. Interactive reasoning benchmarks provide one such direction, offering a controlled setting in which to study how systems learn, adapt, and act in new environments. We present ARC-AGI-3 to serve as a useful platform for advancing research in agentic AI systems and for improving our understanding of what constitutes efficient, general-purpose intelligence.

\section{Acknowledgments}

We thank Surge AI for their early support in brainstorming initial ARC-AGI-3 environment concepts. They provided useful inspiration and helped seed portions of the environment development process.

We also thank the developers who contributed to ARC-AGI-3 for their work in designing and implementing the final set of benchmark environments: Pablo Romero Saavedra, Benjamin Morgan, Vadym Andriianov, Flynn Swainston-Calcutt, Tom Elliot, Fraser Scott, Jonathan Pappas, Kevin Johnson, Lukas Donkers, Danielle Goldman, Majid Manzarpour, Nic Tristani, Mattia Traverso, Christian McDonald, Isaac Karth, Philip Dhingra, and Yago Cerqueira.

\bibliographystyle{plain}
\bibliography{references}

@article{silver2016alphago,
    title = {{Mastering the game of Go with deep neural networks and tree search}},
    author = {David Silver and Aja Huang and Chris J. Maddison and Arthur Guez and Laurent Sifre and George van den Driessche and Julian Schrittwieser and Ioannis Antonoglou and Veda Panneershelvam and Marc Lanctot and Sander Dieleman and Dominik Grewe and John Nham and Nal Kalchbrenner and Ilya Sutskever and Timothy Lillicrap and Madeleine Leach and Koray Kavukcuoglu and Thore Graepel and Demis Hassabis},
    journal = {Nature},
    volume = {529},
    number = {7587},
    pages = {484--489},
    year = {2016},
    doi = {10.1038/nature16961},
    url = {https://doi.org/10.1038/nature16961}
}

@misc{hsu2025quantum,
    title = {{Post on LRM automation discovering novel results in quantum physics}},
    author = {Steve Hsu},
    year = {2025},
    url = {https://x.com/hsu_steve/status/1996034522308026435},
    howpublished = {\url{https://x.com/hsu_steve/status/1996034522308026435}}
}

@misc{gemini3verification,
    title = {{Gemini 3 Deep Think Preview Verification on ARC-AGI-2}},
    author = {{ARC Prize Foundation}},
    year = {2026},
    url = {https://huggingface.co/datasets/arcprize/arc_agi_v2_public_eval/blob/main/gemini-3-deep-think-preview/8698868d.json},
    howpublished = {\url{https://huggingface.co/datasets/arcprize/arc_agi_v2_public_eval}}
}

@misc{arcagi3communityleaderboard,
    title = {{ARC-AGI Community Leaderboard}},
    author = {{ARC Prize Foundation}},
    year = {2026},
    url = {https://github.com/arcprize/ARC-AGI-Community-Leaderboard},
    howpublished = {\url{https://github.com/arcprize/ARC-AGI-Community-Leaderboard}}
}

@misc{smit2025stochasticgoose,
    title = {{ARC3 Solution}},
    author = {Dries Smit},
    year = {2025},
    url = {https://github.com/DriesSmit/ARC3-solution},
    howpublished = {\url{https://github.com/DriesSmit/ARC3-solution}}
}

@misc{blindsquirrel2025,
    title = {{ARC-AGI-3 Agents}},
    author = {wd13ca},
    year = {2025},
    url = {https://github.com/wd13ca/ARC-AGI-3-Agents},
    howpublished = {\url{https://github.com/wd13ca/ARC-AGI-3-Agents}}
}

@misc{vaswani2017attention,
    title = {{Attention Is All You Need}},
    author = {Ashish Vaswani and Noam Shazeer and Niki Parmar and Jakob Uszkoreit and Llion Jones and Aidan N. Gomez and Lukasz Kaiser and Illia Polosukhin},
    year = {2017},
    eprint = {1706.03762},
    archivePrefix = {arXiv},
    primaryClass = {cs.CL},
    url = {https://arxiv.org/abs/1706.03762},
    howpublished = {\url{https://arxiv.org/abs/1706.03762}}
}

@misc{wei2022chainofthought,
    title = {{Chain-of-Thought Prompting Elicits Reasoning in Large Language Models}},
    author = {Jason Wei and Xuezhi Wang and Dale Schuurmans and Maarten Bosma and Brian Ichter and Fei Xia and Ed Chi and Quoc Le and Denny Zhou},
    year = {2022},
    eprint = {2201.11903},
    archivePrefix = {arXiv},
    primaryClass = {cs.CL},
    url = {https://arxiv.org/abs/2201.11903},
    howpublished = {\url{https://arxiv.org/abs/2201.11903}}
}

@misc{chollet2024o3breakthrough,
    title = {{OpenAI o3 Breakthrough High Score on ARC-AGI-Pub}},
    author = {Fran\c{c}ois Chollet},
    year = {2024},
    month = {December},
    url = {https://arcprize.org/blog/oai-o3-pub-breakthrough},
    howpublished = {\url{https://arcprize.org/blog/oai-o3-pub-breakthrough}}
}

@misc{symbolica2026arcgentica,
    title = {{Arcgentica: ARC-AGI-3 Agent Harness Built on the Agentica SDK}},
    author = {Samuel Knutsen and Victoria Klein},
    year = {2026},
    url = {https://github.com/symbolica-ai/ARC-AGI-3-Agents},
    howpublished = {\url{https://github.com/symbolica-ai/ARC-AGI-3-Agents}}
}

@misc{kamradt2025arcagi3preview,
    title = {{ARC-AGI-3 Preview: 30-Day Learnings}},
    author = {Greg Kamradt},
    year = {2025},
    month = {August},
    url = {https://arcprize.org/blog/arc-agi-3-preview-30-day-learnings},
    howpublished = {\url{https://arcprize.org/blog/arc-agi-3-preview-30-day-learnings}}
}

@misc{chollet2019intelligence,
    title={{On the Measure of Intelligence}}, 
    author={Fran\c{c}ois Chollet},
    year={2019},
    eprint={1911.01547},
    archivePrefix={arXiv},
    primaryClass={cs.AI},
    url={https://arxiv.org/abs/1911.01547},
    howpublished = {\url{https://arxiv.org/abs/1911.01547}}
}

@misc{kaggle2020,
    author = {Fran\c{c}ois Chollet and Katherine Tong and Walter Reade and Julia Elliott},
    title = {{Abstraction and Reasoning Challenge}},
    year = {2020},
    howpublished = {\url{https://kaggle.com/competitions/abstraction-and-reasoning-challenge}},
    url = {https://kaggle.com/competitions/abstraction-and-reasoning-challenge},
    note = {Kaggle}
}

@misc{arcathon2023,
    author = {Lab42},
    title = {{ARCathon 2023}},
    year = {2023},
    url = {https://lab42.global/past-challenges/2023-arcathon/},
    howpublished = {\url{https://lab42.global/past-challenges/2023-arcathon/}}
}

@misc{arcathon2022,
    author = {Lab42},
    title = {{ARCathon 2022}},
    year = {2022},
    url = {https://lab42.global/past-challenges/2022-arcathon/},
    howpublished = {\url{https://lab42.global/past-challenges/2022-arcathon/}}
}

@article{spelke2007,
    title={Core knowledge.},
    author={Elizabeth S. Spelke and Katherine D. Kinzler},
    journal={Developmental science},
    year={2007},
    issue={10 1},
    pages={
          89-96
        }
}

@misc{nvarc2025,
    title = {{NVARC Solution to ARC-AGI-2 2025}},
    author = {I. Sorokin and Jean-Francois Puget},
    year = {2025},
    url = {https://drive.google.com/file/d/1vkEluaaJTzaZiJL69TkZovJUkPSDH5Xc/view},
    howpublished = {\url{https://drive.google.com/file/d/1vkEluaaJTzaZiJL69TkZovJUkPSDH5Xc/view}}
}

@misc{fox2026hillclimbing_arcagi3,
    author = {Fox, Alexis and Wang, Junlin and Rosu, Paul and Dhingra, Bhuwan},
    title = {Hill-climbing ARC-AGI-3},
    url = {https://hillclimbarcagi3.notion.site/},
    year = {2026}
}

@misc{arcagi3toolkit,
    title = {{ARC-AGI Toolkit}},
    author = {{ARC Prize Foundation}},
    year = {2026},
    url = {https://github.com/arcprize/ARC-AGI},
    howpublished = {\url{https://github.com/arcprize/ARC-AGI}}
}

@misc{arcprizefoundation,
    title = {{ARC Prize Foundation}},
    year = {2026},
    howpublished = {\url{https://arcprize.org/}},
    url = {https://arcprize.org/},
    note = {Founders: Mike Knoop, Fran\c{c}ois Chollet. Operations: Bryan Landers, Greg Kamradt.}
}

@misc{arcprize2024recap,
    title = {{ARC Prize 2024 Competition}},
    year = {2024},
    url = {https://arcprize.org/competitions/2024},
    howpublished = {\url{https://arcprize.org/competitions/2024}}
}

@misc{arcprize2025recap,
    title = {{ARC Prize 2025 Competition}},
    year = {2025},
    url = {https://arcprize.org/competitions/2025},
    howpublished = {\url{https://arcprize.org/competitions/2025}}
}

@misc{anderson2018evaluation,
      title={On Evaluation of Embodied Navigation Agents}, 
      author={Peter Anderson and Angel Chang and Devendra Singh Chaplot and Alexey Dosovitskiy and Saurabh Gupta and Vladlen Koltun and Jana Kosecka and Jitendra Malik and Roozbeh Mottaghi and Manolis Savva and Amir R. Zamir},
      year={2018},
      eprint={1807.06757},
      archivePrefix={arXiv},
      primaryClass={cs.AI},
      url={https://arxiv.org/abs/1807.06757}, 
}

\end{document}